%% file: NoisyPedagogy.tex
\RequirePackage{amsmath}

\documentclass[graybox]{svmult}

\usepackage{mathptmx}       
\usepackage{helvet}         
\usepackage{courier}        
\usepackage{type1cm}        

\usepackage{makeidx}         
\usepackage{graphicx}        
\usepackage{multicol}        
\usepackage[bottom]{footmisc}

\usepackage{amssymb}
\usepackage{epsfig}
\usepackage{cancel}
\usepackage{color}
\usepackage{booktabs}
\usepackage{enumerate}

\input{h_bib}
\input{h_com}

\makeindex

\begin{document}
\title*{Pragmatic-Pedagogic Value Alignment
}
\author{Jaime~F.~Fisac
	\and Monica~A.~Gates
	\and Jessica~B.~Hamrick
	\and Chang~Liu
    \and Dylan~Hadfield-Menell
    \and Malayandi Palaniappan
    \and Dhruv Malik
	\and S.~Shankar~Sastry
	\and Thomas~L.~Griffiths
	\and Anca~D.~Dragan
 }
\authorrunning{J.~F.~Fisac
	\and M.~A.~Gates
	\and J.~B.~Hamrick
	\and C.~Liu
    \and D.~Hadfield-Menell
    \and et al.
}
\institute{All authors are with the University of California, Berkeley.
\\
\vspace{-0.225in}
\flushleft{\email{{jfisac,mgates,jhamrick,changliu, dhm,malayandi,dhruvmalik,shankar_sastry,tom_griffiths,anca}@berkeley.edu}}
}
\maketitle

\abstract{
As intelligent systems gain autonomy and capability, it becomes vital to ensure that their objectives  match those of their human users; this is known as the value-alignment problem. In robotics, value alignment is key to the design of collaborative robots that can integrate into human workflows, successfully inferring and adapting to their users' objectives as they go. We argue that a meaningful solution to value alignment must combine multi-agent decision theory with rich mathematical models of human cognition, enabling robots to tap into people's natural collaborative capabilities. We present a solution to the cooperative inverse reinforcement learning (CIRL) dynamic game based on well-established cognitive models of decision making and theory of mind. The solution captures a key reciprocity relation: the human will not plan her actions in isolation, but rather reason pedagogically about how the robot might learn from them; the robot, in turn, can anticipate this and interpret the human's actions pragmatically. To our knowledge, this work constitutes the first formal analysis of value alignment grounded in empirically validated cognitive models.
}

\input{1-introduction}
\input{2-theory}

\input{3-results}

\input{4-discussion}
\begin{acknowledgement}
This work is supported by ONR under the Embedded Humans MURI (N00014-13-1-0341), by AFOSR under Implicit Communication (16RT0676), and by the Center for Human-Compatible AI. 
\end{acknowledgement}
\printbibliography
\end{document}

%% file: h_bib.tex
\usepackage{url}
\usepackage[%
	sorting=none,
    backend=biber,
    doi=false,
    isbn=false,
    eprint=false,
    maxcitenames=3, 
    minbibnames=3, 
    maxbibnames=4, 
    giveninits=true,
    block=none]
    {biblatex}

    \renewbibmacro{in:}{}
    \AtEveryBibitem{%
  	\clearlist{language}%
    \clearfield{isbn}%
    \clearfield{pages}%
    \clearfield{editor}%
	}

\DeclareBibliographyCategory{needsurl}
\newcommand{\entryneedsurl}[1]{\addtocategory{needsurl}{#1}}
\renewbibmacro*{url+urldate}{%
  \ifcategory{needsurl}{
    \printfield{url}%
    \iffieldundef{urlyear}
      {}
      {\setunit*{\addspace}%
       \printurldate}}
    {}}

\bibliography{library}
\entryneedsurl{Mitchell2004}

%% file: h_com.tex
\def\E{\mathcal{E}}

\def\I{\mathcal{I}}

\def\EE{\mathbb{E}}

\def\RR{\mathbb{R}}

\newcommand{\aH}{\ensuremath{a_H}}
\newcommand{\aR}{\ensuremath{a_R}}

\newcommand{\bR}{\ensuremath{b_R}}

\usepackage{todonotes}

\usepackage{ifthen}
\newboolean{include-notes}
\setboolean{include-notes}{true}

\newcommand{\adnote}[1]{\ifthenelse{\boolean{include-notes}}%
 {\textcolor{blue}{\textbf{AD: #1}}}{}}
\newcommand{\jfnote}[1]{\ifthenelse{\boolean{include-notes}}%
 {\textcolor{orange}{\textbf{JF: #1}}}{}}
\newcommand{\mgnote}[1]{\ifthenelse{\boolean{include-notes}}%
 {\textcolor{olive}{\textbf{MG: #1}}}{}}
 \newcommand{\jbhnote}[1]{\ifthenelse{\boolean{include-notes}}%
 {\textcolor{purple}{\textbf{JBH: #1}}}{}}
\newcommand{\clnote}[1]{\ifthenelse{\boolean{include-notes}}%
 {\textcolor{cyan}{\textbf{CL: #1}}}{}}
\newcommand{\dhmnote}[1]{\ifthenelse{\boolean{include-notes}}%
 {\textcolor{magenta}{\textbf{DHM: #1}}}{}}
 \newcommand{\tgnote}[1]{\ifthenelse{\boolean{include-notes}}%
 {\textcolor{teal}{\textbf{TG: #1}}}{}}

\newcommand{\adtodo}[1]{\ifthenelse{\boolean{include-notes}}%
 {\todo[author=\textbf{Anca},
 		color=blue!60]{#1}}{}}
\newcommand{\jftodo}[1]{\ifthenelse{\boolean{include-notes}}%
 {\todo[author=\textbf{Jaime},
 		color=orange!60]{#1}}{}}
\newcommand{\mgtodo}[1]{\ifthenelse{\boolean{include-notes}}%
 {\todo[author=\textbf{Monica},
 		color=olive!60]{#1}}{}}
\newcommand{\jbhtodo}[1]{\ifthenelse{\boolean{include-notes}}%
 {\todo[author=\textbf{Jess},
 		color=purple!60]{#1}}{}}
\newcommand{\cltodo}[1]{\ifthenelse{\boolean{include-notes}}%
 {\todo[author=\textbf{Chang},
 		color=cyan!60]{#1}}{}}
\newcommand{\dhmtodo}[1]{\ifthenelse{\boolean{include-notes}}%
 {\todo[author=\textbf{Dylan},
 		color=magenta!60]{#1}}{}}
\newcommand{\tgtodo}[1]{\ifthenelse{\boolean{include-notes}}%
 {\todo[author=\textbf{Tom},
 		color=teal!60]{#1}}{}}

%% file: 1-introduction.tex

\section{Introduction}

The accelerating progress in artificial intelligence (AI) and robotics is bound to have a substantial impact in  society, simultaneously unlocking new potential in augmenting and transcending human capabilities while also posing significant challenges to safe and effective human-robot interaction.
In the short term, integrating robotic systems into human-dominated environments will require them to assess the intentions and preferences of their users in order to assist them effectively, while avoiding failures due to poor coordination.
In the long term, ensuring that advanced and highly autonomous AI systems will be beneficial to individuals and society will hinge on their ability to correctly assimilate human values and objectives \cite{Amodei2017}.
We envision the short- and long-term challenges as being inherently coupled, and predict that improving the ability of robots to understand and coordinate with their human users will inform solutions to the general AI \emph{value-alignment problem}.
 
Successful value alignment requires moving from typical single-agent AI formulations to robots that account for a second {agent---the human---who} determines what the objective is.
In other words, value alignment is fundamentally a multi-agent problem.
Cooperative Inverse Reinforcement Learning (CIRL) formulates value alignment as a two-player game in which a human and a robot share a common reward function, but \emph{only the human} has
knowledge of
this reward \cite{Hadfield-Menell2016a}.
In practice, solving a CIRL game requires more than multi-agent decision theory: we are not dealing with \emph{any} multi-agent system,
but with a human-robot system. 
This poses a unique challenge in that humans do not behave like idealized rational agents \cite{Tversky1974}.
However, humans do
excel at social interaction
and are extremely perceptive of the mental states of others \cite{Heider1944,Meltzoff1995}.
They will naturally project mental states such as beliefs and intentions onto their robotic collaborators, becoming invaluable allies in our robots' quest for value alignment.

In the coming decades, tackling the value-alignment problem will be crucial to building collaborative robots that know what their human users want.
In this paper, we show that value alignment is possible not just in theory, but also in practice.
We introduce a solution for CIRL based on a model of the human agent that is grounded in
cognitive science findings regarding human decision making \cite{Baker2014} and pedagogical reasoning \cite{Shafto2014}.
Our solution leverages two closely related insights to facilitate value alignment.
First,
to the extent that improving their collaborator's understanding of their goals may be conducive to success, people will tend to behave \emph{pedagogically}, deliberately choosing their actions to be informative about these goals.
Second, the robot should anticipate this pedagogical reasoning in interpreting the actions of its human users, akin to how a \emph{pragmatic} listener interprets a speaker's utterance in natural language.
Jointly, pedagogical actions and pragmatic interpretations enable stronger and faster inferences among people \cite{Shafto2014}.
Our result suggests that it is possible for robots to partake in this naturally-emerging equilibrium,
ultimately becoming more perceptive and competent collaborators.

%% file: 2-theory.tex

\section{Solving Value Alignment using Cognitive Models}

\subsection{Cooperative Inverse Reinforcement Learning (CIRL)}
Cooperative Inverse Reinforcement Learning (CIRL) \cite{Hadfield-Menell2016a} formalizes value alignment as a two-player game, which we briefly present here.
Consider two agents, a human $H$ and a robot $R$, engaged in a dynamic
collaborative task involving a (possibly infinite) sequence of steps.
The goal of both agents is to achieve the best possible outcome according to some objective $\theta\in\Theta$.
However, this objective is only known to $H$.
In order to contribute to the objective, $R$ will need to make inferences about $\theta$ from the actions of $H$ (an Inverse Reinforcement Learning (IRL) problem), and $H$ will have an incentive to behave informatively so that $R$ becomes more helpful, hence the term \emph{cooperative} IRL.

Formally, a CIRL game is a dynamic (Markov) game of two players ($H$ and $R$), described by a tuple $\langle S, \{A_H, A_R\}, T, \{\Theta, r\}, P_0, \gamma \rangle$, where
$S$ is the set of possible states of the world;
$A_H,A_R$ are the sets of actions available to $H$ and $R$ respectively;
$T: S \times S \times A_H \times A_R \to [0,1]$ a discrete transition measure%
\footnote{
Note that the theoretical formulation is easily extended to arbitrary measurable sets; we limit our analysis to finite state and objective sets for computational tractability and clarity of exposition.
}
over the next state, conditioned on the previous state and the actions of $H$ and $R$: $T(s'|s,\aH, \aR)$;
$\Theta$ is the set of possible objectives;
$r: S \times A_H \times A_R \times\Theta\to\RR$ is a cumulative reward function assigning a real value to every tuple of state and actions for a given objective: $r(s,\aH,\aR;\theta)$;
$P_0: S\times\Theta\to[0,1]$ is a probability measure on the initial state and the objective;
$\gamma\in[0,1]$ is a geometric time discount factor making future rewards gradually less valuable.

\subsection{Pragmatic Robots for Pedagogic Humans}
Asymmetric information structures in games (even static ones) generally induce an \emph{infinite hierarchy of beliefs}: our robot will need to maintain a Bayesian belief over the human's objectives to decide on its actions.
To reason about the robot's decisions, the human would in principle need to maintain a belief on the robot's belief, which will in turn inform her decisions, thereby requiring the robot to maintain a belief on the human's belief about its own belief, and so on \cite{Zamir2012}.
In \cite{Hadfield-Menell2016a}, it was shown that an \emph{optimal} pair of strategies can be found for any CIRL game by solving a partially observed Markov decision process (POMDP).
This avoids this bottomless recursion as long as both agents are rational and can coordinate perfectly before the start of the game.

Unfortunately, 
when dealing with human agents, 
rationality and prior coordination are nontrivial assumptions.
Finding an equivalent tractability result for more realistic human models is therefore crucial in using the CIRL formulation
to solve real-world value-alignment problems.
We discover the key insight
in cognitive studies of human \emph{pedagogical reasoning} \cite{Shafto2014}, in which a teacher chooses actions or utterances to influence the beliefs of a learner who is aware of the teacher's intention.
The teacher can then exploit the fact that the learner can interpret utterances pragmatically.
Infinite recursion is averted by finding a fixed-point relation between the teacher's best utterance and the learner's best interpretation, exploiting a common modeling assumption in Bayesian theory of mind:
the learner models the teacher as a \emph{noisily rational} decision maker \cite{Luce1959}, who will be \emph{likelier} to choose utterances causing the learner to place a high posterior belief on the correct hypothesis, given the learner's current belief.
While in reality, the teacher cannot exactly compute the learner's belief, the model supposes that she estimates it (from
 the learner's
previous responses to her utterances), then introduces noise in her decisions to capture estimation inaccuracies.
This framework can predict complex behaviors observed in human teaching-learning interactions, in which pedagogical utterances and pragmatic interpretations permit
efficient communication \cite{Shafto2014}.

We adopt an analogous modeling framework to that in \cite{Shafto2014} for value alignment,
with a critical difference: the ultimate objective of the human is not to explicitly improve the robot's understanding of the true objective, but to optimize the team's expected performance \emph{towards} this objective.
Pedagogic behavior thus emerges implicitly to the extent that a well-informed robot becomes a better collaborator.

\subsection{Pragmatic-Pedagogic Equilibrium Solution to CIRL}
The robot does not have access to the true objective $\theta$, but rather estimates a belief $\bR$ over $\theta$.
We assume that this belief on $\theta$ can be expressed parametrically (this is always true if $\Theta$ is a finite set),
and define $\triangle_\Theta$ to be the corresponding (finite-dimensional) parameter space,
denoting $R$'s belief by $\bR\in\triangle_\Theta$.
While in reality the human cannot directly observe $\bR$, we assume, as in~\cite{Shafto2014}, that she can compute it or infer it from the robot's behavior (and model estimation inaccuracies as noise in her policy).
We can then let $Q:S \times \triangle_\Theta \times A_H \times A_R \times\Theta\to\RR$ represent the state-action value function of the CIRL game for a given objective $\theta$, which we are seeking to compute: if $\theta\in\Theta$ is the true objective known to $H$, then $Q(s,\bR,\aH,\aR;\theta)$ represents the best performance the team can expect to achieve if $H$ chooses $\aH$ and $R$ chooses $\aR$ from state $s$, with $R$'s current belief being $\bR$.

In order to solve for $Q$,
we seek to establish an appropriate dynamic programming relation for the game, given a well-defined information structure and a model of the human's decision making.
Since
it is typically possible for people to predict a robot's next action if they see its beginning \cite{Dragan2014},
we assume that $H$ can observe $\aR$ at each turn before committing to $\aH$. 
A well-established model of human decision making in psychology and econometrics is the Luce choice rule, which models people's decisions probabilistically, making high-utility choices more likely than those with lower utility \cite{Luce1959}.
In particular, we employ a common case of the Luce choice rule, the Boltzmann (or \emph{soft-max}) noisy rationality model \cite{Baker2014},
in which the probability of a choice decays exponentially as its utility decreases in comparison to competing options.
The relevant utility metric in our case is the sought $Q$ (which captures $H$'s best expected outcome for each of her available actions $a_H$).
Therefore the probability that $H$ will choose action $a_H$ has the form
\begin{equation}\label{eq:Boltzmann}
  \pi_H^\odot(\aH | s,\bR,\aR; \theta) \propto
  \exp\big(\beta Q(s,\bR,\aH,\aR;\theta)\big)
  \enspace,
\end{equation}
where $\beta>0$ is termed the \emph{rationality coefficient} of $H$ and quantifies the concentration of $H$'s choices around the optimum; as $\beta\to\infty$, $H$ becomes a perfect rational agent, while, as $\beta\to0$, $H$ becomes indifferent to $Q$.
The above expression can be interpreted by $R$ as the \emph{likelihood} of action $\aH$ given a particular $\theta$.
The evolution of $R$'s belief $b_R$ is then given (deterministically) by the Bayesian update
\begin{equation}\label{eq:Bayes}
  b_R'(\theta | s, \bR, \aR,\aH) \propto
  \pi_H^\odot(\aH | s,\bR, \aR; \theta)\bR(\theta)
  \enspace,
\end{equation}
Jointly, \eqref{eq:Boltzmann} and \eqref{eq:Bayes} define a fixed-point equation
analogous to the one in \cite{Shafto2014}, which states how $R$ should pragmatically update $\bR$ based on a noisily rational pedagogic $\aH$. This amounts to a deterministic transition function for $R$'s belief, $\bR' = f_b(s,\bR,\aH,\aR)$.
Crucially, however, the fixed-point relation derived here involves $Q$ itself, which we have yet to compute.

Unlike $H$, $R$ is modeled as a rational agent; however, not knowing the true $\theta$, the best $R$ can do is to maximize%
\footnote{
We assume for simplicity that the optimum is unique or a well-defined disambiguation rule exists.}
the expectation of $Q$ based on its current belief%
\footnote{
Note that this does not imply \emph{certainty equivalence}, nor do we assume separation of estimation and control: $R$ is fully reasoning about how its actions and those of $H$ may affect its future beliefs.
}
$\bR$: 
\begin{equation}\label{eq:robot_policy}
  \pi_R^*(s,\bR) := \arg\max_{\aR}\sum_{\aH,\theta}
  Q(s,\bR,\aH,\aR;\theta)\cdot
  \pi_H^\odot(\aH | s,\bR,\aR; \theta) \bR(\theta)
  \enspace.
\end{equation}

Combining \eqref{eq:Bayes} with the state transition measure $T(s'|s,\aH, \aR)$, we can define the Bellman equation for $H$ under the noisily rational policy $\pi_H^\odot$ for any given $\theta\in\Theta$:
\begin{equation}\label{eq:Q_theta}
  Q(s,\bR,\aH,\aR;\theta)= r(s,\aH,\aR;\theta)\\ +
  \EE_{s',\aH'}\left[
  \gamma\cdot Q'\Big(s',\bR',\aH',\pi_R^*(s',\bR');\theta\Big)\right]
  \;,
\end{equation}
where $s'\sim T(s'|s,\aH, \aR)$;
$\bR'=f_b(s,\bR,\aH,\aR)$;
$\aH'\sim\pi_H^\odot(\aH | s',\bR',\pi_R^*(s',\bR'); \theta)$.
Note that
$H$'s next action $\aH'$ 
implicitly depends on $R$'s action at the next turn.

Substituting (\ref{eq:Boltzmann}-\ref{eq:robot_policy}) into \eqref{eq:Q_theta}, we obtain
the sought dynamic programming relation for the CIRL problem under a noisily rational-pedagogic human and a pragmatic robot.
The human is pedagogic because she takes actions according to (\ref{eq:Boltzmann}), which takes into account how her actions will influence the robot's belief about the objective.
The robot is pragmatic because it assumes the human is actively aware of how her actions convey the objective, and
interprets them accordingly.

The resulting problem is similar to a POMDP (in this case formulated in belief-state MDP form), with the important difference that the belief transition depends on the value function itself. In spite of this complication, the problem can be solved in backward time through dynamic programming: each Bellman update will be based on a pragmatic-pedagogic fixed point that encodes an equilibrium between the $Q$ function (and therefore $H$'s policy for choosing her action) and the belief transition (that is, $R$'s rule for interpreting $H$'s actions).
Evidence in \cite{Shafto2014} suggests that people are proficient at finding such equilibria, even though uniqueness is not guaranteed in general; study of disambiguation is an open research direction.

%% file: 3-results.tex

\section{A Proof-of-Concept}

We introduce the benchmark domain ChefWorld, a household collaboration setting in which a human $H$ seeks to prepare a meal with the help of an intelligent robotic manipulator $R$. There are multiple possible meals that $H$ may want to prepare using the available ingredients, and $R$ does not know beforehand which one she has chosen (we assume $H$ cannot or will not tell $R$ explicitly). The team obtains a reward only if $H$'s intended recipe is successfully cooked. If $H$ is aware of $R$'s uncertainty, she should take actions that give $R$ actionable information, particularly the information that she expects will allow $R$ to be as helpful as possible as the task progresses.

Our problem has 3 ingredients, each with 2 or 3 states: spinach (absent, chopped), tomatoes (absent, chopped, pur\'eed), and bread (absent, sliced, toasted). Recipes correspond to (joint) target states for the food. Soup requires the tomatoes to be chopped then pur\'eed, the bread to be sliced then toasted, and no spinach. Salad requires the spinach and tomatoes to be chopped, and the bread to be sliced then toasted. $H$ and $R$ can slice or chop any of the foods, while only $R$ can pur\'ee tomatoes or toast bread.

\begin{figure}
  \centering
  \includegraphics[width =1.0\linewidth]{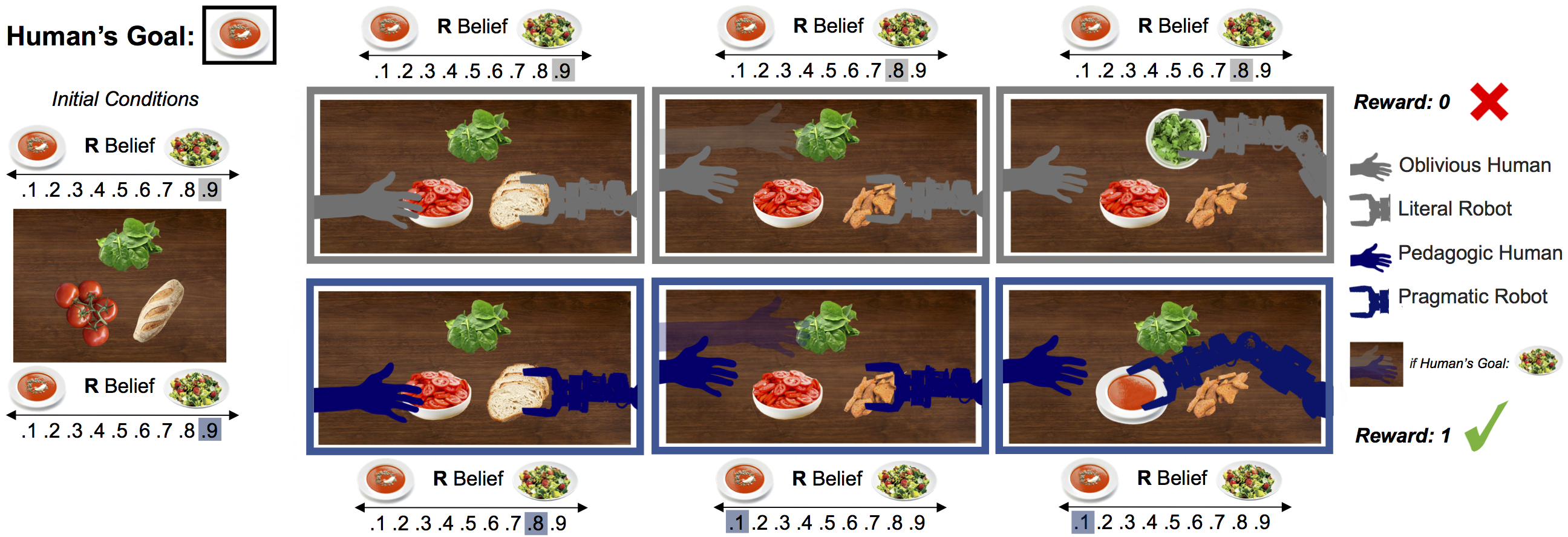}
  \caption{\small
  Simple collaborative scenario with 2 possible objectives.
  The human $H$ wants soup but the robot $R$ initially believes her goal is salad.
  Even under a full POMDP formulation, if $R$ reasons ``literally'' about $H$'s actions using standard IRL 
  (assuming $H$ behaves as if $R$ \emph{knew} the true objective),
  it fails to infer the correct objective.
  Conversely, under the pragmatic-pedagogic CIRL equilibrium, $R$ views $H$ as incentivized to choose pedagogic actions that will fix $R$'s belief when needed.
  Under the pragmatic interpretation,
  $H$'s \emph{wait} action in 
  turn 2 (instead of adding spinach, which would be preferred by a pedagogic $H$ wanting salad)
  indicates $H$ wants soup.
  While $H$'s actions are the same under both solutions, only the pragmatic $R$ achieves value alignment and completes the recipe.
  \label{fig:expt1}  \vspace{-10pt}}
\end{figure}

A simple scenario with the above two recipes is solved using discretized belief-state value iteration and presented as an illustrative example in Fig \ref{fig:expt1}. $R$ has a wrong initial belief about $H$'s intended recipe.
Under standard IRL,
$H$ fails to communicate her recipe. But if $R$ is pragmatic and $H$ is pedagogic, $H$ is able to change $R$'s belief and they successfully collaborate to make the meal.

In addition, we computed the solution to games with 4 recipes through a modification of POMDP value iteration (Table 1).
In the pragmatic-pedagogic CIRL equilibrium with $\beta = 5$, $H$ and $R$ successfully cook the correct recipe 97\% of the time, whereas under the standard IRL framework (with $H$ acting as an expert disregarding $R$'s inferences)
they only succeed 46\% of the time---less than half as often.

\renewcommand{\arraystretch}{1.2}
\begin{table}[h!]
  \centering
  \begin{tabular}{ c|c c c c } 
    \hline
    & \ \ Boltzmann ($\beta$ = 1) &  \ \ Boltzmann ($\beta$ = 2.5) & \ \ Boltzmann ($\beta$ = 5)   &\ \  \qquad Rational\\ 
    \hline
    IRL  & 0.2351 & 0.3783 & 0.4555 & \qquad 0.7083\\
    CIRL  & 0.2916 & 0.7026 & 0.9727 & \qquad 1.0000 \\ 
    \hline
  \end{tabular}
  \vspace{0.1in}
  \caption{A comparison of the expected value (or equivalently here, the probability of success) achieved by CIRL and IRL on the \textit{ChefWorld} domain with four recipes when the robot begins with a uniform belief over the set of recipes. We ran each algorithm across different models of the human's behavior, namely a rational model and a Boltzmann-rational model with various values of $\beta$ (a higher $\beta$ corresponds to a more rational human).
  When the human is highly irrational ($\beta$~=~1), both CIRL and IRL unsurprisingly perform rather poorly.
  However, as the human becomes less noisy ($\beta$~=~2.5, $\beta$~=~5), CIRL outperforms IRL by a significant margin; in fact, the pragmatic-pedagogic CIRL strategy with a Boltzmann-rational human performs comparably ($\beta$~=~2.5) or even substantially outperforms ($\beta$~=~5) the IRL result when the human is perfectly rational.}
  \label{table:1}
\end{table}

%% file: 4-discussion.tex

\section{Discussion}

We have presented here an analysis of the AI value alignment problem that incorporates a well-established model of human decision making and theory of mind into the game-theoretic framework of cooperative inverse reinforcement learning (CIRL). Using this analysis, we derive a Bellman backup that allows solving the dynamic game through dynamic programming. At every instant, the backup rule is based on a pragmatic-pedagogic equilibrium between the robot and the human: the robot is uncertain about the objective and therefore incentivized to learn it from the human, whereas the human has an incentive to help the robot infer the objective so that it can become more helpful.

We note that this type of pragmatic-pedagogic equilibrium, recently studied in the cognitive science literature for human teaching and learning \cite{Shafto2014}, may not be unique in general: there may exist two actions for $H$ and two corresponding interpretations for $R$ leading to different fixed points. For example, $H$ could press a blue or a red button which $R$ could then interpret as asking it to pick up a blue or a red object. Although we might feel that blue-blue/red-red is a more intuitive pairing, blue-red/red-blue is valid as well: that is, if $H$ thinks that $R$ will interpret pressing the blue button as asking for the red object then she will certainly be incentivized to press blue when she wants red; and in this case $R$'s policy should consistently be to pick up the red object upon $H$'s press of the blue button. When multiple conventions are possible, human beings tend to naturally disambiguate between them, converging on salient equilibria or ``focal points'' \cite{schelling1960strategy}. Accounting for this phenomenon is likely to be instrumental for developing competent human-centered robots.

On the other hand, it is important to point out that, although they are computationally simpler than more general multi-agent planning problems, POMDPs are still PSPACE-complete~\cite{mundhenk2000complexity}, so reducing pragmatic-pedagogic equilibrium computation to solving a modified POMDP falls short of rendering the problem tractable in general. However, finding a POMDP-like Bellman backup does open the door to efficient CIRL solution methods that leverage and benefit from the extensive research on practical algorithms for approximate planning in large POMDPs~\cite{silver2010monte}.

We find the results in this work promising for two reasons.
First, they provide insight into how CIRL games can be not only theoretically formulated but also practically solved.
Second, they demonstrate, for the first time, formal solutions to value alignment that depart from the ideal assumption of a rational human agent and instead benefit from modern studies of human cognition.
We predict that developing efficient solution approaches and incorporating more realistic human models will constitute important and fruitful research directions for value alignment.